\documentclass[conference]{IEEEtran}
\IEEEoverridecommandlockouts
\usepackage{cite}
\usepackage{amsmath,amssymb,amsfonts}
\usepackage{algorithmic}
\usepackage{graphicx}
\usepackage{textcomp}
\usepackage{xcolor}
\usepackage{hyperref}
\usepackage{comment}
\usepackage{booktabs}
\usepackage{subfig}
\usepackage{fontawesome5}
\newcommand{\ghfoot}{\href{https://github.com/jedguz/Point–JEPA-grasping}{\faGithub\;\,code \& figures}}

\def\BibTeX{{\rm B\kern-.05em{\sc i\kern-.025em b}\kern-.08em
    T\kern-.1667em\lower.7ex\hbox{E}\kern-.125emX}}
\begin{document}

\title{Label-Efficient Grasp Joint Prediction with Point–JEPA\\}

\author{%
\IEEEauthorblockN{Jed Guzelkabaagac \quad Boris Petrovi\'c}
\IEEEauthorblockA{Technical University of Munich\\
\{jed.guzelkabaagac, boris.petrovic\}@tum.de}
\IEEEauthorblockA{relAI -- Konrad Zuse School of Excellence in Reliable AI}
}

\maketitle
\begingroup
\refstepcounter{footnote}\label{fn:repo}% create a numbered, referenceable footnote
\footnotetext{\ghfoot\ (\url{https://github.com/jedguz/PointJEPA-grasping}).}
\endgroup

\begin{abstract}
We study whether 3D self-supervised pretraining with Point–JEPA enables label-efficient grasp joint-angle prediction. Meshes are sampled to point clouds and tokenized; a ShapeNet-pretrained Point–JEPA encoder feeds a $K{=}5$ multi-hypothesis head trained with winner-takes-all and evaluated by top–logit selection. On a multi-finger hand dataset with strict object-level splits, Point–JEPA improves top–logit RMSE and Coverage@15° in low-label regimes (e.g., \textbf{26\%} lower RMSE at 25\% data) and reaches parity at full supervision, suggesting JEPA-style pretraining is a practical lever for data-efficient grasp learning.
\end{abstract}

\section{Related Work}

Self-supervised learning (SSL) for 3D data has largely progressed along three directions. 

\emph{(i) Reconstruction-based} methods learn by masking and reconstructing inputs in the input space. 
On point clouds this includes point/voxel masked autoencoding; e.g., Voxel-MAE reconstructs masked voxels for sparse automotive LiDAR and improves downstream tasks with fewer labels \cite{Hess_2023,xie2022pointmae,yu2022pointbert,liu2023maskpoint}. 

\emph{(ii) Teacher–student (latent-target)} objectives predict contextualized \emph{latent} targets rather than raw inputs. 
data2vec provides a modality-agnostic recipe that predicts latent representations of the full input from a masked view, while Point2Vec adapts this idea to point clouds and addresses positional-leakage pitfalls unique to 3D, reporting strong transfer on ModelNet40/ScanObjectNN \cite{baevski2022data2vec,zeid2023point2vecselfsupervisedrepresentationlearning}. 

\emph{(iii) Joint-embedding predictive architectures (JEPA)} predict target \emph{representations} for spatially contiguous blocks from a single context block, learning semantic features without heavy view augmentations \cite{ijepa,hu20243djepajointembeddingpredictive}. 
Point–JEPA brings this design to point clouds with a simple sequencer that orders patch centers so contiguous context/target blocks can be sampled despite unordered points, avoiding input-space reconstruction or extra modalities while remaining competitive on standard 3D benchmarks \cite{saito2025pointjepajointembeddingpredictive,hu20243djepajointembeddingpredictive}.

Despite this progress, the impact of predictive, context-aware pretraining on \emph{grasp–joint prediction} remains underexplored \cite{mahler2017dexnet,gualtieri2016gpd,sundermeyer2021contactgraspnet}. 
In this study, we adopt a Point–JEPA backbone for object embeddings and pair it with an inference-aware, multi-hypothesis loss for joint angles, comparing against training from scratch and analyzing label-efficiency gains under fixed object-level splits.

\begin{figure}[!t]
  \centering
  \subfloat[]{\includegraphics[width=0.32\columnwidth]{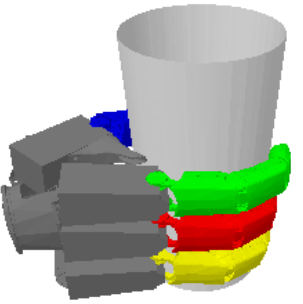}}\hfill
  \subfloat[]{\includegraphics[width=0.32\columnwidth]{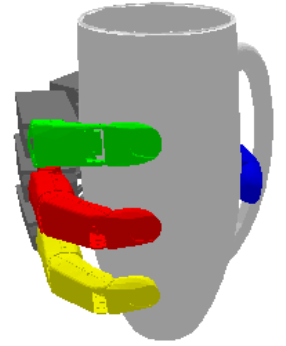}}\hfill
  \subfloat[]{\includegraphics[width=0.32\columnwidth]{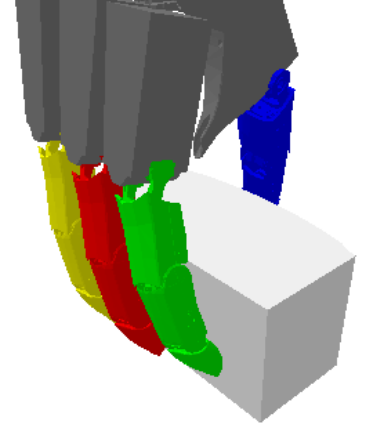}}
  \caption{\textbf{Qualitative predictions.} Conditioned on an object mesh and a given 6-DoF hand pose, the model predicts a \emph{12D} joint-angle vector that yields a stable, 
  collision-free grasp. Panels (a–c) show the \emph{top-logit} hypothesis ($K{=}5$) for three object–pose pairs.}
  \label{fig:grasp-examples}
  \vspace{-2.5mm}
\end{figure}

\begin{figure*}[!t]
  \centering
  \includegraphics[width=0.92\textwidth]{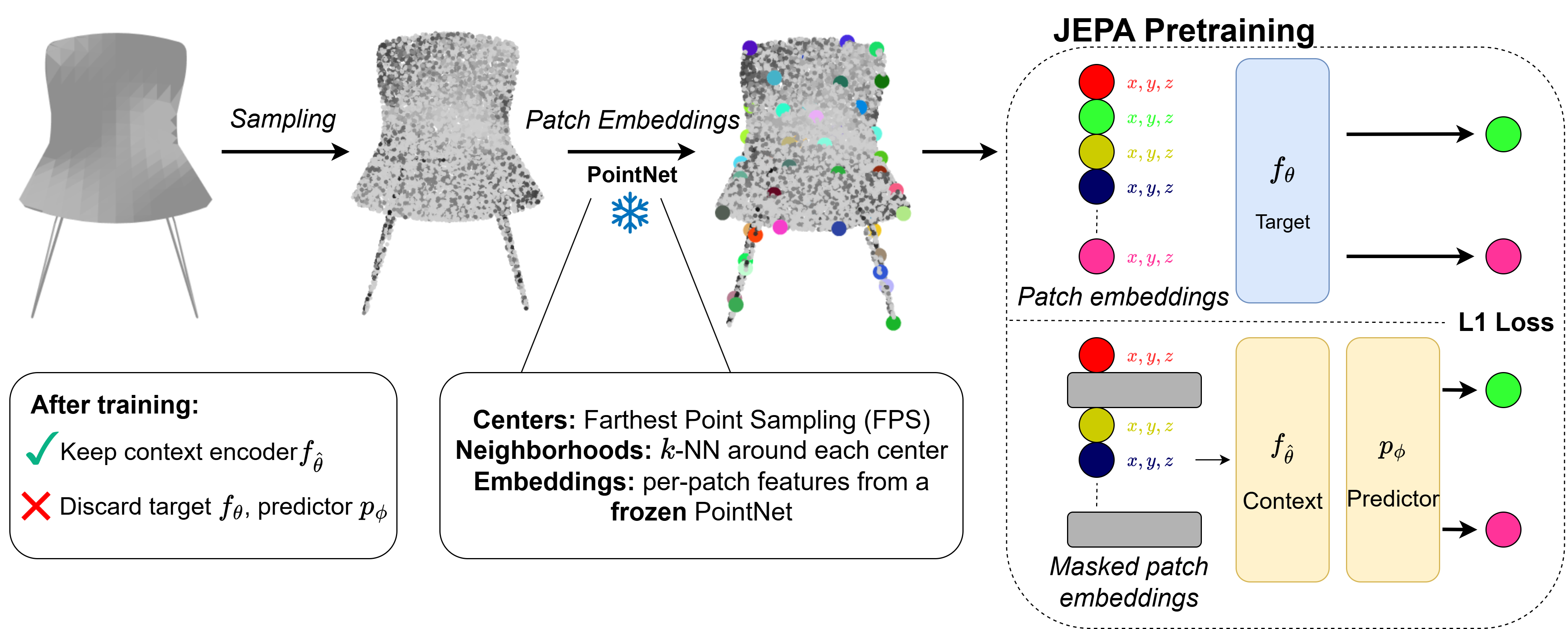}
  \vspace{-1mm}
\caption{\textbf{Pretraining pipeline.}
Meshes $\rightarrow$ point clouds; FPS+$k$-NN patches embedded by a \emph{frozen} PointNet; JEPA learns contextualized patch features \cite{saito2025pointjepajointembeddingpredictive}.
\emph{Schematic by the authors. We fine-tune a backbone initialized from the official Point–JEPA checkpoint; JEPA pretraining is from \cite{saito2025pointjepajointembeddingpredictive} and uses ShapeNet only.}}
  \label{fig:pipeline}
  \vspace{-1mm}
\end{figure*}

\section{Technical Outline}

Meshes are sampled into point clouds and grouped into local patches for transformer processing \cite{qi2017pointnetplusplus}.
Using the Point–JEPA encoder pretrained on ShapeNet \cite{saito2025pointjepajointembeddingpredictive},
we follow the JEPA masking scheme \cite{ijepa}: a block sampler picks spatially contiguous target windows and
a simple sequencer orders patch centers so those windows remain contiguous in the token sequence.
The context encoder sees masked tokens; the target encoder (an EMA copy) sees the full input and provides
stop-gradient latent targets. A lightweight predictor maps context features to the target space.
After pretraining we discard the predictor and target encoders and retain the context encoder as the backbone.
Attention pooling yields a single object embedding (Fig.~\ref{fig:pipeline}).

For grasp prediction, the object embedding is concatenated with the 6-DoF hand pose
(3D translation + unit quaternion; 7 parameters) and passed to a lightweight head that emits $K$
joint-angle hypotheses and logits. Training uses a winner-takes-all / min-over-$K$ objective to preserve
multi-modality (Sec.~\ref{sec:head-loss}); at inference, we select the \emph{top–logit} hypothesis \cite{phanminh2020covernet}.

\section{Methodology}\label{sec:methodology}

\subsection{Dataset \& split protocol}

\noindent\textbf{Data and filtering.}
We use a non-public grasp dataset \cite{9981133}, accessed under educational use, built atop ShapeNet object categories, comprising a range from 10 to 500 unique shapes per category. Each shape provides around 240 grasp samples, each consisting of a 6-DoF hand pose, a 12D joint configuration, and a grasp quality score. Following common practice, we discard grasps with score \(<1.5\), yielding on the order of 1{,}500 meshes and 370{,}000 total grasps.

\medskip
\noindent\textbf{Object-level, category-stratified splits.}
To prevent overfitting and pose leakage from multiple grasps of the same mesh, we split at the object level. Sample-level splits proved overly easy and invalidated the value of SSL-based encoding.

\medskip
\noindent\textbf{Fixed evaluation.}
We fix one suite consistent for all label budgets: \emph{val} (10\% of objects) and \emph{test} (another 10\%).

\medskip
\noindent\textbf{Low–data training subsets.}
For $p\in\{1, 10, 25, 100\}\%$, we sample the train set from the remaining objects with proportional \emph{category stratification} (at least one object per synset).

\medskip
\noindent\textbf{Splits \& reproducibility.}
We use a fixed, object-level split pack (Pack~A) for all label budgets. To assess robustness, we built an independent Pack~B and verified at 25\% and 100\% that trends mirror Pack~A. For reproducibility, we release code that \emph{deterministically regenerates} both packs from a local dataset root (see Footnote~\ref{fn:repo}); we \emph{do not publish} split JSONs or any third-party file paths/assets.

\subsection{Preprocessing \& object representation}

Meshes are converted to point clouds and tokenized into local groups to create
patch embeddings. We use $1024$ points per object and a tokenizer
with $64$ groups (group size $32$, radius $0.05$). These patch embeddings feed
the self–supervised encoder (Fig.~\ref{fig:pipeline}); a global object embedding
is then produced by \emph{attention pooling}, which we found to be more stable
and accurate than mean/max pooling.

\subsection{SSL backbone (Point–JEPA)}
We use the Point–JEPA \emph{context encoder} as the backbone (predictor/target discarded after pretraining). Tokenized point-cloud patches with positional encodings are encoded and attention-pooled into a single object embedding. We fine-tune end-to-end with a \emph{two-tier learning rate} (smaller on the pretrained backbone, larger on the new grasp head); concrete settings and schedules appear in Sec.~\ref{sec:results}.

\subsection{Joint estimation head \& loss}\label{sec:head-loss}

We predict $K$ candidate joint configurations $\{\hat{\mathbf{j}}_{k}\}_{k=1}^{K}$ for each object embedding and hand pose, together with a logit vector $\boldsymbol{\ell}\!\in\!\mathbb{R}^{K}$ used to rank hypotheses at test time. Grasping is inherently \emph{multi-modal}: several distinct joint settings can succeed for the same pose, and single-output MSE collapses these modes to their mean, often yielding an infeasible grasp.

To preserve distinct modes, we adopt the standard winner–takes–all (WTA; min-over-$K$) objective—widely used in multimodal regression and structured prediction \cite{10.5555/2999325.2999336, rupprecht2017learninguncertainworldrepresenting}—and previously applied to dexterous grasping \cite{humt2023combiningshapecompletiongrasp}. For completeness, the training objective is:

\begin{align}
k^* &= \arg\min_k \|\hat{\mathbf{j}}_{k}-\mathbf{j}\|^{2} \\
L   &= \|\hat{\mathbf{j}}_{k^*}-\mathbf{j}\|^{2}
      + \alpha\,\mathrm{CE}(\boldsymbol{\ell},\,k^*)
\label{eq:min-k}
\end{align}

The first term penalizes only the closest hypothesis (avoiding mean collapse); the cross-entropy term trains the logits to act as a selector for the winning mode. At inference, we use the \emph{same} selector (pick the top–logit hypothesis), avoiding any oracle “best-of-$K$” evaluation. This matches common multi-hypothesis practice in trajectory forecasting and related settings.

We briefly evaluated Mixture Density Networks (MDNs) as a probabilistic alternative, but observed instability in this high-dimensional joint space and poorer alignment with our top-1 test-time selection; the WTA formulation proved simpler and more stable in practice.

\begin{figure}[!t]
  \centering
  \includegraphics[width=\columnwidth]{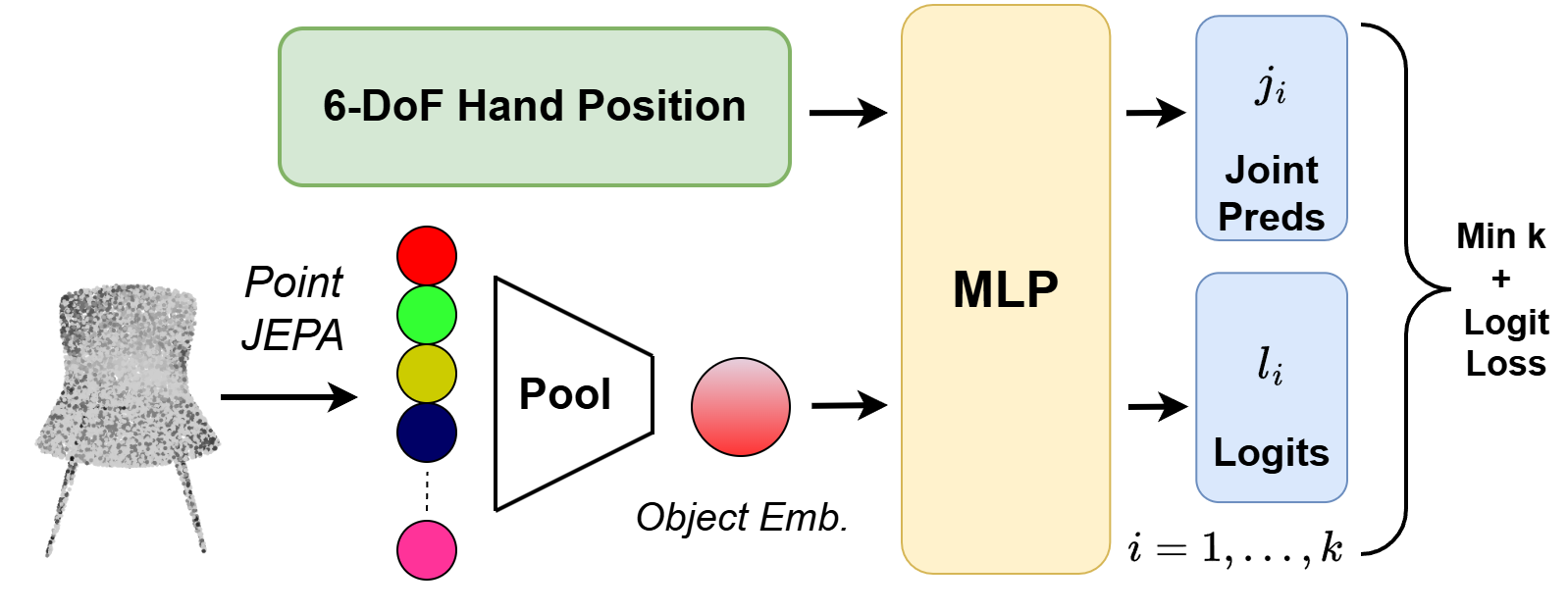}
  \caption{\textbf{Grasp head architecture.}
  Contextualized object features from Point–JEPA are attention-pooled to a global embedding and concatenated with the 6-DoF hand pose. An MLP predicts $K$ joint configurations and corresponding logits.}
  \label{fig:grasp-head}
  \vspace{-1mm}
\end{figure}

\subsection{Evaluation protocol \& metrics}

At inference we select the \emph{top–logit} hypothesis. The primary metric is \textbf{top–logit RMSE} on joint angles.
We also report \textbf{Coverage@$15^\circ$} (a sample is covered if any hypothesis is within $15^\circ$ of ground truth) and the \textbf{selection gap}
$\Delta_{\text{sel}}=\mathrm{RMSE}_{\text{top–logit}}-\mathrm{RMSE}_{\text{best-of-}K}$ (lower is better).
Unless stated otherwise, $K{=}5$.

\section{Results}\label{sec:results}

All results use object-level stratification on the validation split and are averaged over seeds (and split packs where available).

\subsection{Main outcomes}
Table~\ref{tab:toplogit_final} reports top–logit RMSE (rad; lower is better).
\begin{itemize}
  \item \textbf{1–10\% labels:} consistent gains ($\approx$8–10\% relative), indicating improved label efficiency.
  \item \textbf{25\% labels:} largest improvement ($\approx$26\% relative), a “sweet spot” before full supervision saturates.
  \item \textbf{100\% labels:} parity within seed variance (pretraining has little impact compared to scratch).
\end{itemize}

\subsection{Inference-aware selection and coverage}
To assess whether the logits reliably choose the correct hypothesis, we define the \emph{selection gap}
\[
\Delta_{\text{sel}} \;=\; \mathrm{RMSE}_{\text{top–logit}} \;-\; \mathrm{RMSE}_{\text{best-of-}K},
\]
with smaller values indicating better alignment between the learned selector and the oracle winner.
At 1\% and 10\% budgets, JEPA reduces the gap relative to scratch (e.g., 1\%: $0.142$ vs $0.165$~rad; 10\%: $0.157$ vs $0.176$~rad), demonstrating stronger inference-aware selection. The full curves and plotting scripts are provided in our Git repository.

\begin{table}[t]
\centering
\caption{\textbf{Top–logit RMSE (rad) across label budgets.}
Means $\pm$ SD over seeds; budgets marked {\dag} average across Packs A+B (25\%, 100\%), while 1\% and 10\% use Pack~A only.}
\label{tab:toplogit_final}
\begin{tabular}{lccc}
\toprule
Train split & Scratch & JEPA & $\Delta$ (rel.)\\
\midrule
 1\%   & 0.363 $\pm$ 0.002 & \textbf{0.335 $\pm$ 0.003} & +7.7\% \\
 10\%  & 0.335 $\pm$ 0.003 & \textbf{0.303 $\pm$ 0.009} & +9.6\% \\
 25\%{\dag}  & 0.332 $\pm$ 0.002 & \textbf{0.246 $\pm$ 0.012} & +25.9\% \\
 100\%{\dag} & 0.235 $\pm$ 0.002 & \textbf{0.234 $\pm$ 0.008} & +0.4\% \\
\bottomrule
\end{tabular}
\end{table}

\begin{comment}
\begin{figure}[b!]
  \centering
  \subfloat[]{\includegraphics[width=0.3\columnwidth]{grasp_mink_1.png}}\hspace{0.3cm}
  \subfloat[]{\includegraphics[width=0.3\columnwidth]{grasp_mink_2.png}}
  \caption{\textbf{Min-over-$K$ illustrated.} For a fixed object and wrist pose, two hypotheses $(\hat{\mathbf{j}}_{1},\hat{\mathbf{j}}_{2})$ yield stable, collision-free grasps (a–b), while their mean is infeasible. Training (Eq.~\ref{eq:min-k}) regresses only the hypothesis nearest to $\mathbf{j}$ by L2 in joint-angle space and trains logits via cross-entropy; inference uses the top–logit.}
  \label{fig:min_k_predictions}
\end{figure}
\end{comment}

As a complementary multi-hypothesis metric, we report Coverage@$15^\circ$: a grasp is covered if \emph{any} head lies within $15^\circ$ of the ground-truth joint angles.
Coverage increases with label budget and is consistently higher with JEPA in the low-label settings (e.g., at 10\%: $0.955$ vs $0.938$; at 1\%: $0.866$ vs $0.861$); see the repository for the corresponding curves and scripts.

\subsection{Sensitivity to learning rates (10\% grid)}
We ran a 3$\times$2 grid over $(\mathrm{LR}_{\mathrm{backbone}},\,\mathrm{LR}_{\mathrm{head}})$ on the 10\% split (time constraints precluded repeating per budget).
Top–logit RMSE versus training step shows a broad plateau at convergence: final differences across the grid are small ($\lesssim 0.01$\,rad).
Larger backbone LRs destabilize early fine-tuning, while smaller head LRs slow convergence.
We therefore fix $(\mathrm{LR}_{\mathrm{backbone}},\,\mathrm{LR}_{\mathrm{head}})=(1{\times}10^{-5},\,1{\times}10^{-3})$ for all budgets.

\section{Conclusion}
We investigated whether 3D self-supervised pretraining improves grasp–joint prediction under limited labels. 
Integrating a Point–JEPA backbone with a winner-takes-all multi-hypothesis head yields consistent gains in low-label regimes on the multi-finger hand dataset, with the largest relative improvement at 25\% data. 
At 100\% labels, training from scratch attains parity. 
Diagnostics indicate that the selector is effective (smaller selection gap) and that Coverage@$15^\circ$ improves with pretraining, supporting the inference-aware design.

Results indicate JEPA-style pretraining is a practical label-efficiency lever for grasping systems when annotation is expensive.

\section{Future work}
We outline several directions to strengthen generalization and deployment relevance:
\begin{itemize}
  \item \textbf{Cross-domain evaluation.} Run the pre-registered \textit{test\_category} suite and real-robot tests to assess transfer beyond object-level splits.
  \item \textbf{Heads and selection.} Sweep $K$ and study diversity-encouraging variants, while monitoring selection gap and top–logit coverage as primary inference-aware metrics.
\item \textbf{Geometry-aware patching.} Replace FPS\,+\,fixed-kNN with patches built from simple geometric cues, e.g., multi-scale radius neighborhoods, curvature-guided grouping, and small overlaps, so each patch better respects local shape.
\item \textbf{Geometry-focused tokenizer.} Swap the PointNet patch encoder for one that uses explicit local geometry. The JEPA objective stays the same; only the patch encoder changes.
\item \textbf{Compute-efficient fine-tuning.} Evaluate lightweight adaptation (e.g., LoRA) and backbone LR schedules across label budgets.
\end{itemize}

\begin{figure}[t!]
  \centering
  \includegraphics[width=\linewidth]{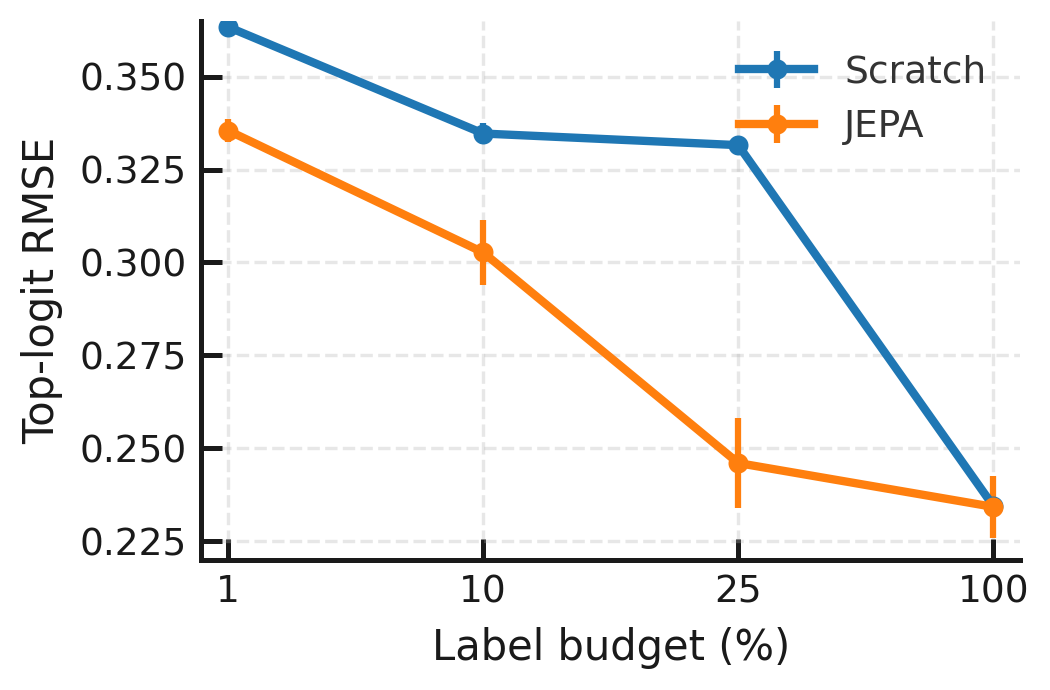}
  \caption{\textbf{Label efficiency.}   Val top–logit RMSE versus label budget (1\%, 10\%, 25\%, 100\%).
    JEPA pretraining improves performance in low-label regimes, with the largest gap at 25\%; at 100\% both methods converge.
    Points show mean over seeds/split packs; error bars denote $\pm$1\,SD.}
  \label{fig:validation_RMSE_1}
  \vspace{-2mm}
\end{figure}

\section*{Acknowledgment}
This work is supported by the DAAD programme Konrad Zuse Schools of Excellence in Artificial Intelligence (relAI), sponsored by the Federal Ministry of Research, Technology and Space (BMFTR).

\bibliographystyle{IEEEtran}
\bibliography{refs}
\vspace{12pt}

\end{document}